# Kinematic and stiffness analysis of the Orthoglide, a PKM with simple, regular workspace and homogeneous performances

Anatoly Pashkevich, Philippe WENGER, Damien CHABLAT

Robotic Laboratory, Department of Control Syatems,
Belarusian State University of Informatics and Radioelectronics
6 P.Brovka St., Minsk 220027, Belarus
pap@bsuir.unibel.by
Institut de Recherche en Communications et Cybernétique de Nantes UMR CNRS 6597
1, rue de la Noe, BP 92101, 44321 Nantes Cedex 03 France
Philippe.Wenger@irccyn.ec-nantes.fr, Damien.Chablat@irccyn.ec-nantes.fr

*Abstract*— **The Orthoglide is a Delta-type PKM dedicated to 3-axis rapid machining applications that was originally developed at IRCCyN in 2000-2001 to meet the advantages of both serial 3-axis machines (regular workspace and homogeneous performances) and parallel kinematic architectures (good dynamic performances and stiffness). This machine has three fixed parallel linear joints that are mounted orthogonally. The geometric parameters of the Orthoglide were defined as function of the size of a prescribed cubic Cartesian workspace that is free of singularities and internal collision. The interesting features of the Orthoglide are a regular Cartesian workspace shape, uniform performances in all directions and good compactness. In this paper, a new method is proposed to analyze the stiffness of overconstrained Delta-type manipulators, such as the Orthoglide. The Orthoglide is then benchmarked according to geometric, kinematic and stiffness criteria: workspace to footprint ratio, velocity and force transmission factors, sensitivity to geometric errors, torsional stiffness and translational stiffness.**

## I. INTRODUCTION

The question whether a parallel-kinematic machine (PKM) is globally more suitable for rapid machining than a serial machine or not, is difficult to answer [1, 2] and still open. PKMs and serial machines have their own merits and drawbacks. Today, most industrial 3-axis machine-tools have a serial kinematic architecture with orthogonal linear joint axes along the x, y and z directions. Thus, the motion of the tool in any of these directions is linearly related to the motion of one of the three actuated axes. Also, the performances are constant throughout the Cartesian workspace, which is a parallelepiped. The main drawback is inherent to the serial arrangement of the links, namely, poor dynamic performances. The Orthoglide is a translational 3-axis Delta-type [3, 4] PKM that was designed to meet the advantages of serial machine tools but without their drawbacks [3]. Starting from a Delta-type architecture with three fixed linear joints and three articulated parallelograms, an optimization procedure was conducted on the basis of two kinematic criteria (i) the conditioning of the Jacobian matrix of the PKM and (ii) the transmission factors. The first criterion was used to impose an isotropic configuration where the tool forces and velocities are equal in all directions. The second criterion was used to define the actuated joint limits and the link lengths with respect to a desired Cartesian workspace size and prescribed limits on the transmission factors. The Orthoglide has a Cartesian workspace shape that is close to a cube whose sides are parallel to the planes xy, yz and xz respectively.

In this paper, a new method is proposed to analyze the stiffness of overconstrained Delta-type manipulators, such as the Orthoglide. The Orthoglide is then benchmarked according to geometric, kinematic and stiffness criteria: workspace to footprint ratio, velocity and force transmission factors, sensitivity to geometric errors, torsional stiffness and translational stiffness. Next section recalls the geometric and kinematic parameters of the Orthoglide prototype. The main contribution compared to our previous results [1], [3], [5] – [8] is related to the Orthoglide stiffness analysis, which is based on the developed analytical pseudo-rigid model. This allows comparing the Orthoglide to the similar performances of other manipulators [9, 10].

## II. GEOMETRY AND KINEMATICS OF THE ORTHOGLIDE

Figure. 1 shows a photography (left) and a CAD-model (right) of the Orthoglide prototype. This machine has three parallel *PRPaR* identical chains (where *P*, *R* and *Pa* stand respectively for prismatic, revolute, and parallelogram joints). The actuated joints are the three orthogonal prismatic ones. The output body is connected to the prismatic joints through a set of three parallelograms, so that it can move only in translation. Note that because only two parallelograms would be sufficient to restrict the motion in translation, the Orthoglide is overconstrained. The Orthoglide has been designed so that it has an isotropic configuration in its workspace, that is, a configuration where the Jacobian matrix is isotropic. This configuration is reached when all parallelograms are orthogonal to each other (Fig. 2). To have a kinematic behavior close to the one of a serial 3-axis machine tool, we have also imposed that, in the isotropic configuration, the velocity transmission factors must be equal to 1.

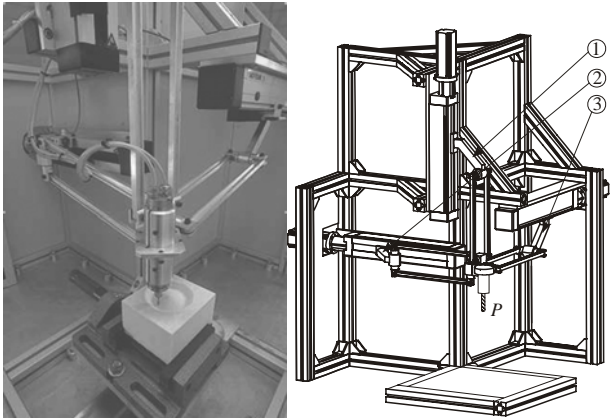

Figure 1: The Orthoglide prototype (©CNRS Photothèque/CARLSON Leif)

This condition implies that for each leg, the axis of the linear joint and the axis of the parallelogram are collinear. Since at the isotropic configuration, the parallelograms are orthogonal, this implies that the linear joints are orthogonal (Fig. 2). An optimization scheme was developed to calculate automatically the geometric parameters as function of the size $L_{workspace}$ of a prescribed Cartesian workspace, which is a cube [3]. The three main steps of this scheme are briefly recalled.

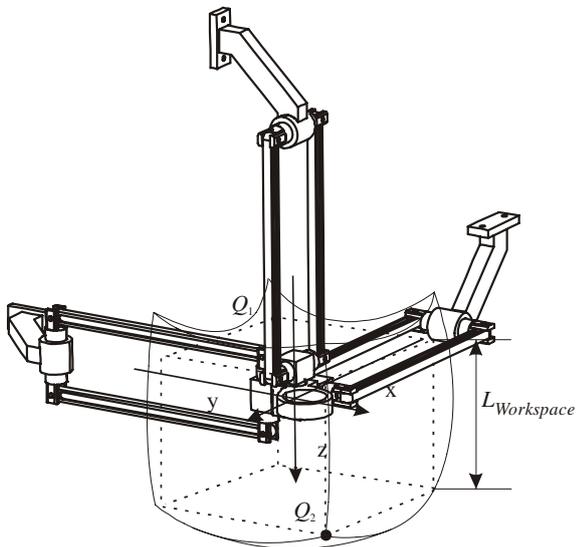

Figure 2: Isotropic configuration and Cartesian workspace of the Orthoglide mechanism and points $Q_1$ and $Q_2$.

First, two points $Q_1$ and $Q_2$ are determined in the prescribed cubic Cartesian workspace (Fig. 2) such that if the transmission factor bounds are satisfied at these points, they are satisfied in all the prescribed Cartesian workspace. These points are then used to define the leg length as function of $L_{workspace}$. Finally, the positions of the base points and the linear actuator range are calculated such that the prescribed cubic Cartesian workspace is fully included in the Cartesian workspace of the Orthoglide. The prototype built at IRCCyN was designed using this optimization scheme on the basis of the following data: the size of the prescribed workspace is $L_{workspace}$ = 200 mm and the minimal and maximal velocity transmission factors λ are ½ and 2, respectively as shown in Fig. 3 [7]. These factors are defined as the eigenvalues of the product of the Jacobian matrix **J** by its transpose The resulting length of the three parallelograms is 310 mm and the resulting range of the linear joints is 257 mm. The values of the transmission factors inside the prescribed cubic workspace were confirmed using interval analysis [8]. Fig. 4 shows a view of the orthoglide workspace computed with interval analysis, in which we could verify the inclusion of the prescribed cubic workspace.

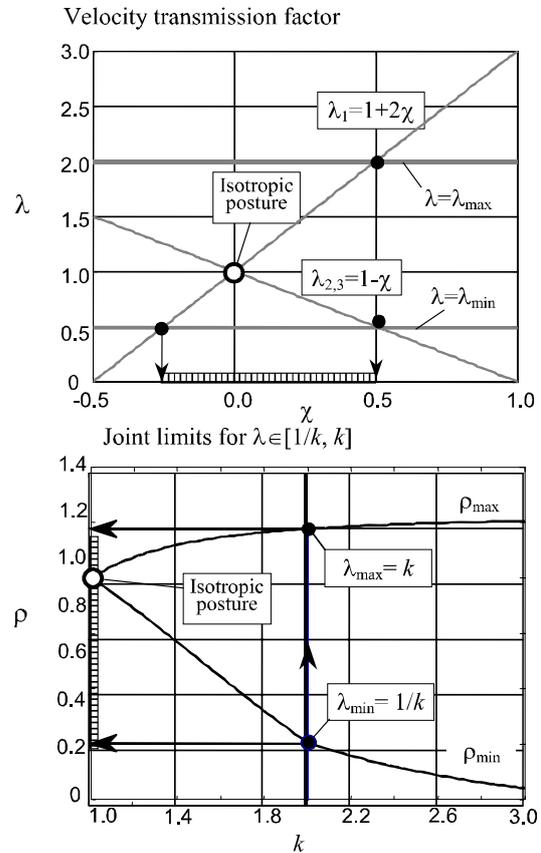

Figure 3: Computing the joint limits of the velocity transmission factors along $Q_1$ and $Q_2$

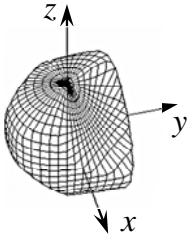

Figure 4 Workspace calculated with interval analysis for $\lambda \in [1/2, 2]$

### III. SENSIBILITY ANALYSIS

Two complementary methods were used to analyze the sensitivity of the Orthoglide to its dimensional and angular variations [5]. The first method was used to have a rough idea of the influence of the dimensional variations on the location of the end-effector. It showed that variations in the design parameters of the same type from one leg to the other have the same influence on the end-effector (Fig. 5). The second method takes into account the variations in the parallelograms. It uses a differential vector method to study the influence of the dimensional and angular variations in the parts of the manipulator on the position and orientation of the end-effector, and particularly the influence of the variations in the parallelograms.

It turns out that the kinematic isotropic configuration of the manipulator is the least sensitive one to its dimensional and angular variations. On the contrary, the closest configurations to its kinematic singular configurations are the most sensitive ones to geometrical variations.

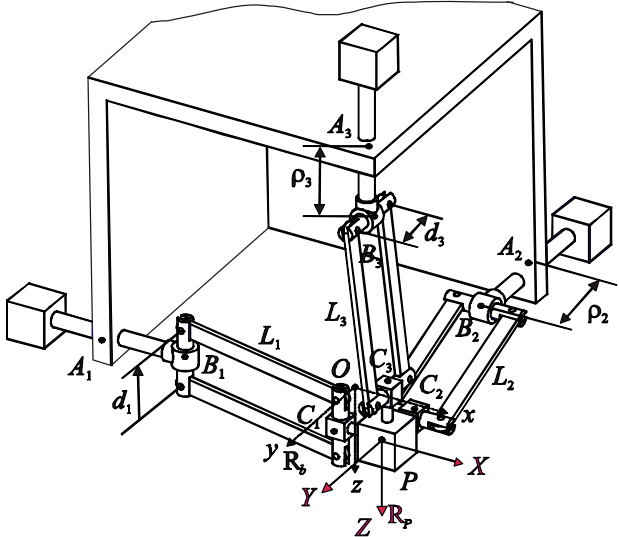

Figure 5: Basic kinematic architecture of the Orthoglide

The position of the end-effector is sensitive to variations in the lengths of the parallelograms, $\Delta L_i$, and to the position errors of points $A_i$, $B_i$, and $C_i$ along axis $x_i$. Conversely, the influence of $\Delta l_i$ and $\Delta m_i$, the parallelism errors of the parallelograms, is low and even negligible in the kinematic isotropic configuration. The orientation errors of the prismatic joints are the most influential angular errors on the position of the end-effector. Besides, the position of the end-effector is not sensitive to angular variations in the isotropic configuration. $\Delta l_i$ and $\Delta m_i$ are the only dimensional variations, which are responsible for the orientation error of the end-effector. However, the influence of the parallelism error of the small sides of the parallelograms, depicted by $\Delta l_i$, is more important than the one of the parallelism error of their long sides, depicted by $\Delta m_i$. The sensitivity of the position and the orientation of the end-effector is generally null in the kinematic isotropic configuration, and is a maximum when the manipulator is close to a kinematic singular configuration, i.e. $P = Q_2$.

Indeed, only two kinds of design parameters variations are responsible for the variations in the position of the end-effector in the isotropic configuration: $\Delta L_i$ and $\Delta e_{ix}$. Likewise, two kinds of variations are responsible for the variations in its orientation in this configuration: $\Delta l_i$, the parallelism error of the small sides of the parallelograms, $\Delta \theta_{Aiy}$ and $\Delta \gamma_{iy}$. Moreover, the sensitivities of the pose (position and orientation) of the end-effector to these variations are a minimum in this configuration, except for $\Delta_i$. On the contrary, $Q_2$ configuration is the most sensitive configuration of the manipulator to variations in its design parameters. Variations in the pose of the end-effector depend on all the design parameters variations and are a maximum in this configuration.

We are able to compute the variations in the position and orientation of the end-effector with knowledge of the amount of variations in the design parameters. For instance, let us assume that the parallelism error of the small sides of the parallelograms is equal to 10 μm, then, the position error of the end-effector will be equal about to 3μm in $Q_1$ configuration. Likewise, if the orientation error of the direction of the i[th] prismatic joint about axis $y_i$ of $R_i$ is equal to 1 degree, the position error of the end-effector will be equal about to 4.8 mm in $Q_2$ configuration.

Table 1 shows that the probability to get a position error lower than 0.3 mm is higher in the kinematic isotropic configuration than in $Q_1$ and $Q_2$ configurations, if we assume that the length and angular tolerances are equal to 0.05 mm and 0.03 deg, respectively. However, the probability to get an orientation error lower than 0.25 deg is the same in $Q_1$ and the isotropic configurations, but is lower in $Q_2$ configuration.

TABLE 1. PROBABILITIES TO GET A POSITION ERROR LOWER THAN 0.3 MM AND AN ORIENTATION ERROR LOWER THAN 0.25 DEG IN $Q1$, $Q2$ AND ISOTROPIC CONFIGURATIONS

|  | Configuration | | |
| --- | --- | --- | --- |
|  | $Q_1$ | Isotropic | $Q_2$ |
| $\text{Prob}(\|\delta_P\|_2 \leq 0.3mm)$ | 0.8468 | 0.9683 | 0.7276 |
| $\text{Prob}(\delta\theta \leq 0.25\deg)$ | 0.9691 | 0.9690 | 0.9453 |

## IV. STIFFNESS ANALYSIS

For Parallel Kinematic Machines, stiffness is an essential performance measure since it is directly related to the positioning accuracy and the payload capability. Mathematically, this benchmark is defined by the stiffness matrix, which describes the relation between the linear/angular displacements of the end-effector and the external forces/torques applied to the tool. At present there are three main methods for the computation of the stiffness matrix: the Finite Element Analysis (FEA) [11], the matrix structural analysis (SMA) [12], and the virtual joint method (VJM) that is often called the lumped modeling [13].

The first of them, FEA, is proved to be the most accurate and reliable, however it is usually applied at the final design stage because of the high computational expenses required for the repeated re-meshing of the complicated 3D structure over the whole workspace. The second method, SMA, also incorporates the main ideas of the FEA, but operates with rather large elements – 3D flexible beams describing the manipulator structure. This obviously leads to the reduction of the computational expenses, but does not provide with clear physical relations required for the parametric stiffness analysis. And finally, the VJM method is based on the expansion of the traditional rigid model by adding the virtual joints (localized springs), which describe the elastic deformations of the links. The VJM technique is widely used at the pre-design stage and is also convenient for the PKM benchmarking.

Let us apply the VJM method to evaluate the stiffness of the Orthoglide. However, in contrast to the previous work [14], let us consider the model, which does not include auxiliary revolute joints that are usually added to ensure feasibility of the traditional VJM technique. Thus, each leg includes an actuated joint $\rho$ and three passive joints $q_1$, $q_2$, $q_3$ presented in Fig. 6 (assuming that the parallelogram posture is defined by a single joint variable $q_2$). The flexibility of each leg is modeled by seven virtual joints $\theta_0 \ldots \theta_6$ presented in Table 2 where $L_f$, $b_f$, $h_f$ are the geometrical parameters of the foot (length and cross-sectional dimensions), $L_B$, $S_B$ and $d$ are the parallelogram geometrical parameters, $E$ and $G$ are Young's & Coulomb's modulus of the link material, and $I_{f1}$, $I_{f2}$, $I_0$ are the quadratic and polar moments for the corresponding cross-sections. Under this assumption, the differential kinematic model of each leg can be presented as

$$\delta t = J_i^{(\theta)} \cdot \delta \theta_i + J_i^{(q)} \cdot \delta q_i \qquad (1)$$

where the superscript $i \in \{x, y, z\}$ refers to the particular manipulator leg, $\delta\theta_i = [\delta\theta_{0i}, \ldots, \delta\theta_{6i}]^T$ is vector of the *reactive* joint coordinates (which respond to the external force), $\delta q_i = [\delta q_{1i}, \ldots, \delta q_{3i},]^T$ is the vector of the *passive* joint coordinates (which do not respond to the force), $\delta t = [\delta p_x, \delta p_y, \delta p_z, \delta\varphi_x, \delta\varphi_y, \delta\varphi_z]^T$ is the deviation of the end-platform location (which includes both the position $p_x$, $p_y$, $p_z$ and orientation $\varphi_x$, $\varphi_y$, $\varphi_z$), and $J_\theta$ and $J_q$ are the Jacobians of the size 6×7 and 6×3 respectively.

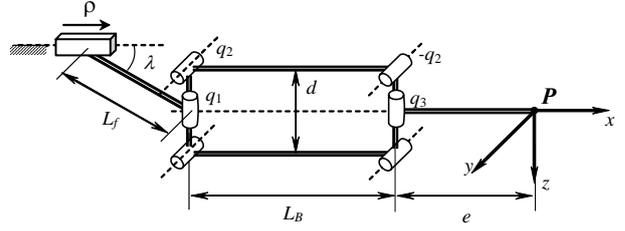

Figure 6: Geometry of the Orthoglide leg and location of the passive joints

The static equilibrium condition gives supplementary relations, which can be derived using the principle of virtual work. According to it, the external force $f_i$ applied to the leg end-point satisfies the following equations

$$J_i^{(\theta)^T} f_i = K_i^{(\theta)} \delta\theta_i; \quad J_i^{(q)^T} f_i = 0 \qquad (2)$$

where $K_i^{(\theta)} = diag(k_{0i}, \ldots k_{6i})$ is the stiffness matrix of the reactive joints. Hence, after elimination of $\delta\theta_I$, the combined kinematic/static leg model can be written as

$$\begin{bmatrix} \mathbf{S}_i^{(\theta)} & \mathbf{J}_i^{(q)} \\ \mathbf{J}_i^{(q)^T} & \mathbf{0} \end{bmatrix}_{9\times 9} \cdot \begin{bmatrix} \mathbf{f}_i \\ \delta \mathbf{q}_i \end{bmatrix}_{9\times 1} = \begin{bmatrix} \delta \mathbf{t} \\ \mathbf{0} \end{bmatrix}_{9\times 1}. \quad (3)$$

where $\mathbf{S}_i^{(\theta)} = \mathbf{J}_i^{(\theta)} \mathbf{K}_i^{(\theta)^{-1}} \mathbf{J}_i^{(\theta)^T}$ is the matrix of the size of 6×6 defining influence of the reactive joints at the end-effector. From (3), the leg stiffness may be computed by direct inversion of relevant 9×9 matrix and selection of its diagonal part of size 6×6. However, the most reliable solution is based on the SVD-decomposition of the Jacobian $\mathbf{J}_i^{(q)} = \mathbf{U}_i^{(q)} \cdot \mathbf{\Sigma}_i^{(q)} \cdot \mathbf{V}_i^{(q)^T}$, which yields the following expression:

$$\mathbf{K}_i = \mathbf{U}_{id}^{(q)} \; (\mathbf{U}_{id}^{(q)^T} \; \mathbf{S}_i^{(\theta)} \; \mathbf{U}_{id}^{(q)})^{-1} \; \mathbf{U}_{id}^{(q)^T} \quad (4)$$

where $\mathbf{U}_{di}^{(q)}$ is the 6×3 matrix, which is obtained by the partitioning of the orthogonal matrix $\mathbf{U}_i^{(q)} = [\mathbf{U}_{ri}^{(q)} \; \mathbf{U}_{di}^{(q)}]$ into two parts corresponding to the non-zero and zero singular values respectively.

TABLE 2. VIRTUAL JOINTS AND THEIR STIFFNESS PARAMETERS

| No | Feature | Stiffness | Figure |
|---|---|---|---|
| $\theta_0$ | Actuator stiffness | $k_0 = k_{act}$ | |
| $\theta_1$ | Foot bending due to force $F$ | $k_1 = 3EI_{f1}/L_f$ | |
| $\theta_2$ | Foot bending due to torque $T$ | $k_2 = 2EI_{f2}/L_f$ | |
| $\theta_3$ | Foot tension due to torque $T$ | $k_3 = GI_{f0}/L_f$ | |
| $\theta_4$ | Foot rotation due to torque $T$ | $k_4 = EI_{f2}/L_f$ | |
| $\theta_5$ | Parallelogram tension due to force $F$ | $k_5 = 2ES_b/L_b$ | |

| $\theta_6$ | Parallelogram tension due to torque $T$ | $k_5 = ES_b d^2 \cos^2$ | 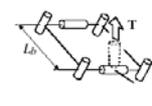 |

Taking into account all kinematic chains (i.e. legs X, Y, Z) and applying the superposition principle, the stiffness matrix of the whole manipulator structure can be expressed as

$$\mathbf{K} = \sum_{i \in \{x,y,z\}} \mathbf{K}_i$$

Compared to our previous results on the Orthoglide stiffness analysis [14], the obtained expressions gives more clear physical meaning of the leg stiffness matrix. In particular, it includes the matrix product, which is usual for the manipulators without passive joints. Obviously, the latter is the primary source of the manipulator reaction to the external force/torque. But this reaction is slackened by the passive joints taken into account by the matrix $U_d$, which eliminates force/torque components in the directions $u_1, \ldots u_r$ that are accepted by the passive joints. Apparently, for the case without passive joints, this expression is reduced to the known one. It should be also noted that for a single manipulator leg, the stiffness matrix K is the singular (since the leg is able to respond to end-platform displacements in the directions $u_{r+1}, \ldots u_6$ only). For instance, the rotation around the additional joint axis $q_4$ used in [14] is completely accepted by this joint without any contribution of the remaining ones (both reactive and passive). The latter is reflected in the matrix **K** by having the zero row and the zero column, but these column/row locations within the matrix are different for the Orthoglide legs X, Y, Z.

For the entire manipulator, the stiffness matrix is the non-singular within the dextrous workspace. It is because the union of the "reactive" subspaces of three legs (defined by the linear spans of the corresponding vectors $u_{r+1}, \ldots u_6$) completely covers the entire 6-dimentional space of the force/torque. For instance, the rotation around the X-axis, which is non-reactive for the X-leg because of the passive joint q4, causes the reaction of the remaining legs Y and Z. Hence, for the non-singular manipulator postures (when $r = 4$), each leg produces an independent 2-dimensional reactive subspace, which in the conjunction yield the full force/torque space of the dimension 6.

Using the derived expression for the matrix **K**, we computed the stiffness benchmarks for the Orthoglide prototype described in Section 2. As follows from relevant results presented in Table 3, within the dexterous workspace of size 200×200×200 mm$^3$ the translational and rotational stiffness varies by 30…40% only, which is a good result for parallel manipulators that usually posses non-isotropic rigidity.

The model validity has been evaluated using the FEA method. The obtained results demonstrated rather high accuracy (10…15%) allowing easy integration of the developed stiffness model in the design optimizations routines. However, experimental study produced higher errors, which are caused by low stiffness of the prototype base frame. This motivates further research, which must take into account additional sources of the stiffness.

TABLE 3. STIFFNESS OF THE ORTHOGLIDE PROTOTYPE

| Translational stiffness [N/mm] | Rotational stiffness [N·mm/rad] |
|---|---|
| *Isotropic point* ($x, y, z = 0$) ||
| $\begin{bmatrix} 2.71 & 0 & 0 \\ 0 & 2.71 & 0 \\ 0 & 0 & 2.71 \end{bmatrix} \cdot 10^3$ | $\begin{bmatrix} 8.37 & 0 & 0 \\ 0 & 8.37 & 0 \\ 0 & 0 & 8.37 \end{bmatrix} \cdot 10^6$ |
| *Point $Q_1$* ($x, y, z = -73.65$ mm) ||
| $\begin{bmatrix} 1.65 & -0.65 & -0.65 \\ -0.65 & 1.65 & -0.65 \\ -0.65 & -0.65 & 1.65 \end{bmatrix} \cdot 10^3$ | $\begin{bmatrix} 8.95 & 1.16 & 1.16 \\ 1.16 & 8.95 & 1.16 \\ 1.16 & 1.16 & 8.95 \end{bmatrix} \cdot 10^6$ |
| *Point $Q_2$* ($x, y, z = +126.35$ mm) ||
| $\begin{bmatrix} 2.61 & 2.18 & 2.18 \\ 2.18 & 2.61 & 2.18 \\ 2.18 & 2.18 & 2.61 \end{bmatrix} \cdot 10^3$ | $\begin{bmatrix} 11.07 & 0.09 & 0.09 \\ 0.09 & 11.07 & 0.09 \\ 0.09 & 0.09 & 11.07 \end{bmatrix} \cdot 10^6$ |

Another benchmark is based on the tool displacement for the standard groove milling operations along the *y*-axis (detailed technical data on the manufacturing conditions are given in [15]). The corresponding external forces and torques are equal to:

$$\mathbf{F} = [\ F_x,\ F_y,\ F_z,\ -F_y\ h_z,\ F_x\ h_z,\ 0\ ]^T$$

where

$F_x = 215N$; $F_y = -10N$; $F_z = -25N$ and $h_z = 100$ mm

The displacement vector $\delta\mathbf{t}$ at the point $P$ can be found from the expression

$$\delta\mathbf{t} = \mathbf{K}\cdot\mathbf{F}$$

with

$$\delta\mathbf{t} = [\delta p_x,\ \delta p_y,\ \delta p_z,\ \delta\varphi_x,\ \delta\varphi_y,\ \delta\varphi_z]^T.$$

However, because of the tool offset $\mathbf{h} = [h_x\ h_y\ h_y]^T$, the displacement at the point TCP (Tool Centre Point) must be adjusted using the expression

$$\delta\mathbf{t}_{TCP} = \begin{bmatrix} 0 & -\delta\varphi_z & \delta\varphi_y & \delta p_x \\ \delta\varphi_z & 0 & -\delta\varphi_x & \delta p_y \\ -\delta\varphi_y & \delta\varphi_x & 0 & \delta p_z \\ 0 & 0 & 0 & 1 \end{bmatrix} \cdot \begin{bmatrix} h_x \\ h_y \\ h_z \\ 1 \end{bmatrix}$$

which for $h_x = h_y = 0$ yields

$$\delta p_x^{TCP} = \delta p_x + h_z\,\delta\varphi_y$$
$$\delta p_y^{TCP} = \delta p_y - h_z\,\delta\varphi_x$$
$$\delta p_z^{TCP} = \delta p_z$$

Numerical results for the examined case study are presented in Table 4. As follows from them, the largest position error is achieved in the *x*-axis direction, it is rather high and is equal to 0.67 mm and 0.35 mm for the non-constrained and over-constrained models respectively. These results show that the values of the design parameters that could not be optimized by the kinematic criteria, namely link sections, must be larger.

TABLE 4. POSITION AND ORIENTATION ERRORS FOR THE MILLING OPERATION (*ISOTROPIC CONFIGURATION*)

| | $\delta p_x$, mm | $\delta p_y$, mm | $\delta p_z$, mm | $\delta\varphi_x$, rad | $\delta\varphi_y$, rad | $\delta\varphi_z$, rad |
|---|---|---|---|---|---|---|
| P-point | 0.0792 | -0.0037 | -0.0092 | -0.0003 | 0.0027 | -0.0004 |
| TCP | 0.3482 | 0.0239 | -0.0092 | -0.0003 | 0.0027 | -0.0004 |

## V. CONCLUSIONS

This paper analyzed kinematic and stiffness properties of the Orthoglide, a 3-axis Delta-type, overconstrained PKM that was designed and optimized to meet the performances of both serial and parallel machines. A new method was developed to analyse the translational and rotational stiffness, taking into account the overconstrained model. The Orthoglide was benchmarked against workspace to footprint ratio, velocity and force transmission factors, sensitivity to geometric errors, and translational and rotational stiffness. The isotropic design with bounded transmission factors, which was originally conducted to provide a regular workspace shape with homogeneous force and velocities distribution throughout the workspace, also proved to lead to stable stiffness and sensitivity performances.

ACKNOWLEDGMENTS

This work was partially supported by the European project NEXT, acronyms for "Next Generation of Productions Systems", Project no° IP 011815.

REFERENCES

[1] J. Tlusty, J. Ziegert and S. Ridgeway, "Fundamental Comparison of the Use of Serial and Parallel Kinematics for Machine Tools," Annals of the CIRP, Vol.48:1, pp.351-356, 1999.
[2] P. Wenger, C. Gosselin, and B. Maille, "A comparative study of serial and parallel mechanism topologies for machine tools," in *Proc. PKM'99*, Milan, Italy, 1999, pp. 23–32.
[3] D. Chablat and P. Wenger., "Architecture Optimization of a 3-DOF Parallel Mechanism for Machining Applications, the Orthoglide," IEEE Transactions On Robotics and Automation, Vol. 19/3, pp. 403-410, June 2003.
[4] R. Clavel. Delta, "A fast robot with parallel geometry," In 18th Int. Symp. On Industrial Robots, pages 91–100, Lausanne, April 1988. IFS Publications.
[5] S. Caro, P. Wenger, F. Bennis and Chablat D., "Sensitivity Analysis of the Orthoglide, a 3-DOF Translational Parallel Kinematic Machine," ASME Journal of Mechanical Design, Vol. 128, pp. 392-402, March 2006.
[6] A. Pashkevich D. Chablat and P. Wenger, "Kinematics and Workspace Analysis of a Three-Axis Parallel Manipulator: the Orthoglide," Robotica, january 2006.
[7] A. Pashkevich, P. Wenger and D. Chablat, "Design Strategies for the Geometric Synthesis of Orthoglide-type Mechanisms," Journal of Mechanism and Machine Theory, Volume 40, Issue 8, pp. 907-930, August 2005.
[8] D. Chablat, P. Wenger, F. Majou and J-P Merlet, "An Interval Analysis Based Study for the Design and the Comparison of 3-DOF Parallel Kinematic Machines," International Journal of Robotics Research, 2004.
[9] O. Company, F. Pierrot, and J.C. Fauroux, "A Method for Modeling Analytical Stiffness of a Lower Mobility Parallel Manipulator," In: IEEE International Conference on Robotics and Automation (ICRA), pp. 3243–3248, Barcelona, Spain, April 2005.
[10] R. Rizk, N. Andreff, J.C. Fauroux, J.M. Lavest and G. Gogu, "Precision study of a decoupled four degrees of freedom parallel robot including manufacturing and assembling errors," In: 6th International Conference on Integrated Design and Manufacturing in Mechanical Engineering (IDMME 2006), 12 pp, Grenoble, France, May 2006, 12p (CD-ROM Proceedings).
[11] B.C. Bouzgarrou, J.C. Fauroux, G. Gogu and Y. Heerah, "Rigidity analysis of T3R1 parallel robot uncoupled kinematics," in: Proc. of the 35th International Symposium on Robotics (ISR), Paris, France, March 2004, 6 p.
[12] D. Deblaise, X.Hernot and P.Maurine, "A Systematic Analytical Method for PKM Stiffness Matrix Calculation," In: *IEEE International Conference on Robotics and Automation (ICRA)*, pp.4213-4219, Orlando, Florida - May 2006.
[13] C.M. Gosselin, "Stiffness mapping for parallel manipulators," *IEEE Transactions on Robotics and Automation*, Vol. 6, 1990, pp. 377–382.
[14] F. Majou, C. Gosselin, P. Wenger, D. Chablat. "Parametric stiffness analysis of the Orthoglide," Mechanism and Machine Theory, 2006, in press.
[15] F. Majou, C. Gosselin, P. Wengerand D. Chablat, "Parametric Stiffness Analysis of the Orhtoglide," International Symposium on Robotics, 2004.